\documentclass[letterpaper]{article}

\usepackage[utf8]{inputenc}
\usepackage{CJKutf8}      
\usepackage[T1]{fontenc}
\usepackage{times}          
\usepackage{helvet}         
\usepackage{courier}        
\usepackage{amsmath, amssymb, amsfonts, mathtools, amsthm}

\usepackage[pdftex]{graphicx}
\usepackage{grffile}        
\usepackage{tikz}
\usetikzlibrary{shapes.misc, positioning}
\usepackage{subfig}         
\usepackage{floatrow}       
\usepackage[section]{placeins}

\usepackage{booktabs}       
\usepackage{multirow}
\usepackage{colortbl}

\usepackage{microtype}      
\usepackage[none]{hyphenat} 
\usepackage{parskip}        
\usepackage{xcolor}         
\usepackage[normalem]{ulem} 

\usepackage{tcolorbox}
\tcbuselibrary{breakable}   
\usepackage{listings}       
\usepackage{algpseudocode}  

\usepackage[numbers]{natbib}
\usepackage{url}
\usepackage{hyperref}
\hypersetup{
    colorlinks  = true,
    linkcolor   = purple,
    citecolor   = teal,
    urlcolor    = teal,
    linktoc     = all
}

\usepackage{authblk}        
\usepackage{thmtools, thm-restate}
\usepackage{wrapfig}
\usepackage{nicefrac}
\usepackage{macros}         
\usepackage{basic}          
\usepackage{enumitem}       
\usepackage[pro]{fontawesome5} 

\title{TCMQA: A 38K-Question Traditional Chinese Medicine Benchmark with a Licensed-Practitioner Reference}

\author[$\dagger$,$*$]{Tzu-Heng~Huang}
\author[$\ddagger$,$*$]{Jet~Lin}
\author[$\S$]{Eric~Lin}

\affil[$\dagger$]{University of Wisconsin-Madison}
\affil[$\ddagger$]{University of California, Merced}
\affil[$\S$]{TechTCM}

\begin{document}

\maketitle

{\renewcommand{\thefootnote}{\fnsymbol{footnote}}%
\footnotetext[1]{Equal contribution. Contact through: \url{thuang273@wisc.edu} \& \url{jetlin101@gmail.com}}}

\begin{abstract}
Medical benchmarks for language models are built almost entirely on Western biomedicine.
Traditional Chinese Medicine (TCM) is a separate system, with its own diagnostic framework and its own literature, and it remains largely unmeasured.
The few TCM evaluations that exist are small, narrow, and rarely paired with a human reference.
We present \textsc{TCMQA}, an open benchmark of 38{,}279 questions from Chinese TCM licensing examinations, paired with 15{,}151 responses from 101 licensed practitioners.
We evaluate 29 instruction-tuned models from 9 families, spanning 0.27B to 14.8B parameters.
Accuracy ranges over 59 points, and no model approaches saturation.
Pretraining data predicts TCM ability far better than scale: a 12B Western-pretrained model reaches $39.6\%$, while a Chinese-pretrained model an eighth its size reaches $60.8\%$.
Nine models exceed the practitioner majority vote of $64.9\%$, the best by $21.8$ points, and all nine come from that same Chinese-pretrained family.
Yet difficulty does not transfer between models and practitioners: accuracy is flat across practitioner-rated difficulty, item-level agreement is near zero for all 29 models, and on $8.4\%$ of items the practitioners are correct where the leading model is wrong.
We release the corpus, the practitioner responses, the harness, and per-item model outputs at~\url{https://huggingface.co/datasets/TechTCM/TCMQA}.
\end{abstract}
\section{Introduction}
\label{introduction}

%
Large language models (LLMs) now perform competitively on open-ended question answering, complex reasoning, and specialized tasks in physics, law, and medicine.
These models are now part of daily medical use: patients ask them for diagnostic suggestions and help reading prescriptions, and clinicians use them for case analysis and documentation~\cite{bean2026reliability, meng2024application, kunze2025large, mingole2025dr, jindal2024large, roustan2025clinicians, busch2025current}.
A large body of medical benchmarks has followed, evaluating factual knowledge, safety, and clinical reasoning~\cite{wu2026medarena, pal2022medmcqa, jin2021disease, jin2019pubmedqa}.
Most of this work assumes Western biomedicine, whether the questions are posed in English or translated into other languages.
Traditional Chinese Medicine (TCM), a separate medical system with its own diagnostic framework and its own therapeutic principles, remains largely unmeasured.

%
Normal users bear most of the cost of this gap.
Patients now ask general-purpose assistants which herb suits a presenting pattern, whether two formulas conflict, and what a prescription written by their clinician means~\cite{liu2025evaluating, han2026tuning, li2025llm}.
They ask in Chinese, and the models answer fluently in Chinese whether or not they know any TCM.
Fluency, however, says nothing about whether an answer is grounded in TCM, and few leaderboards report TCM ability, so users have no practical way to check.
Privacy concerns push many toward open-weight models they can run locally---often the smaller models least likely to have absorbed TCM knowledge during pretraining.

%
Existing TCM evaluations do not close this gap either~\cite{yue2024tcmbench, cheng2025tcm}.
They are typically small---a few thousand items---narrow in coverage, and either closed or only partly released.
Each of these limits what a score can tell us: a small test set cannot separate models that finish close together, thin coverage leaves whole areas of the field untested, and a closed set cannot be reproduced.
Nor do these benchmarks report a human reference point.
Accuracy is hard to interpret in isolation: the natural comparison is what licensed practitioners score on the same items.

%
We introduce \textsc{TCMQA}, an open benchmark built from Chinese TCM licensing examinations, with three components.
The \emph{corpus} holds 38{,}279 multiple-choice items in a uniform five-option format, covering the professional curriculum from materia medica and formula composition through classical texts and professional regulation; 2{,}386 of them have more than one correct option.
The \emph{human reference} comes from 101 licensed practitioners, who contributed 15{,}151 responses through an annotation platform we built.
On 4{,}801 items three practitioners responded independently, and those items form the subset for every human--model comparison.
Practitioners also supplied domain labels and a difficulty rating recorded before the answer was scored.
The \emph{harness} scores each model in a single pass, reading a prediction both from the generated text and from the first-token log-probabilities.
We apply it to 29 instruction-tuned models from 9 families, ranging from 0.27B to 14.8B parameters and spanning Western- and Chinese-developed systems.

%
The benchmark separates the field by 59 accuracy points, from $18.95\%$ to $77.98\%$, and no model saturates it.
We find that pretraining composition matters, not scale: within a family accuracy rises monotonically with size, but across families size predicts little.
Gemma-3-12B-It~\cite{kamath2025gemma} reaches $39.60\%$, while Qwen2-1.5B-Inst.~\cite{qwen2024qwen25}, with an eighth of the parameters, reaches $60.81\%$; the OLMo family~\cite{olmo2025olmo, olmo20252} never clears $27\%$ at any size we tested.
Nine models exceed the practitioner majority vote of $64.94\%$, the strongest by $21.79$ points, and all nine descend from a single Chinese-pretrained family.

%
Moreover, practitioners agree substantially about which items are hard, yet model accuracy is flat across that consensus: on the 616 items all three practitioners missed, the model leading the subset scores $87.34\%$, slightly \emph{above} its $86.73\%$ average on the subset as a whole.
Their difficulty ratings say the same thing: practitioner accuracy falls 57 points from the easiest level to the hardest, while most model lines barely move.
Models are furthest ahead on Regulation area, which rewards recall, and closest on diagnosis and pattern differentiation, which needs clinical reasoning.

%
We make the following contributions.
\begin{itemize}[leftmargin=1.5em, itemsep=0.25em, topsep=0.35em]
    \item \textbf{An open TCM benchmark at scale.} 38{,}279 items in a uniform five-option format, single- and multi-answer, released in full---an order of magnitude larger than prior TCM evaluations.
    \item \textbf{A licensed-practitioner reference.} 15{,}151 responses from 101 practitioners, 4{,}801 items with three independent annotations each, plus domain labels, difficulty ratings, and quality flags---to our knowledge the first for TCM.
    \item \textbf{A reproducible testbed.} A fixed single-pass protocol that scores each model from both its generated text and its log-probabilities, applied to 29 instruction-tuned models from 9 families.
\end{itemize}

\section{Dataset and Evaluation Testbed}
\label{framework}

%
\textsc{TCMQA} comprises three components: (i) a corpus of 38{,}279 examination items, (ii) a human reference collected from licensed practitioners through an in-house annotation platform, and (iii) an automated harness that evaluates models under a fixed protocol.
We describe each in turn.

\subsection{Dataset Construction}
\label{sec:dataset}

%
Our corpus is assembled from two sources of Chinese TCM licensing examination material: an in-house collection of practice examinations used to prepare candidates for licensure, and a web-crawled archive of historical licensing papers.
Much of this material exists only in PDF form.
We process it with \textsc{Qwen2.5-14B-Instruct}, prompting the model to read each document and convert each item into a JSON record under a fixed schema: the question, five labeled options, the reference answer, and any explanation or difficulty value the source supplied.
Extracted records then pass through a schema filter: we keep an item only if it has exactly five options and a reference answer.
Our human study also serves as an independent audit, as practitioners could flag any item as defective or incorrect, and we exclude flagged items from the final results.

\begin{table}[t]
\centering
\caption{
The ten-domain TCM taxonomy, with the domain portfolio of the 4{,}801-item comparison subset.
\emph{Items} is the number of subset items assigned the domain by at least one of their three raters, and \emph{Share} is that count as a fraction of 4{,}801.
}
\label{tab:domains}
\small
\begin{tabular}{c p{3.3cm} r r p{5.4cm}}
\toprule
\# & Domain & Items & Share & Scope \\
\midrule
1  & Foundational Theory \newline \zh{中醫基礎理論}   & 1{,}304 & 27.2\% & Yin--Yang, Five Phases, visceral manifestation, qi and blood \\
2  & Classical Texts \newline \zh{中醫經典文獻}       &    700 & 14.6\% & The canonical medical literature and its interpretive tradition \\
3  & Diagnosis \& Patterns \newline \zh{診斷與辨證學} & 1{,}309 & 27.3\% & The four examinations; pattern differentiation and treatment \\
4  & Meridians \& Acupuncture \newline \zh{經絡與針灸學} &  502 & 10.5\% & Channel theory, point location, needling and moxibustion \\
5  & Materia Medica \newline \zh{中藥本草學}          & 1{,}494 & 31.1\% & Herb properties, formula composition, preparation, contraindication \\
6  & Internal Medicine \newline \zh{中醫內科學}       & 1{,}982 & 41.3\% & Internal disorders, incl.\ gynecology, pediatrics, mental illness \\
7  & External Medicine \newline \zh{中醫外科學}       &    698 & 14.5\% & Surface disorders, dermatology, traumatology \\
8  & ENT \& Ophthalmology \newline \zh{中醫五官科學}  &     84 &  1.7\% & Disorders of the eye, ear, nose, and throat \\
9  & Regulation \newline \zh{中醫法規}                &    495 & 10.3\% & Licensure, practice scope, drug control, professional ethics \\
10 & Other \newline \zh{其他}                         & 1{,}419 & 29.6\% & Items not covered above \\
\bottomrule
\end{tabular}
\end{table}

\begin{table}[t]
\centering
\caption{Composition of the \textsc{TCMQA} corpus.}
\label{tab:corpus}
\begin{tabular}{lrr}
\toprule
Property & Count & Share \\
\midrule
Total items                       & 38{,}279 & 100.0\% \\
\quad single-answer               & 35{,}893 & 93.8\% \\
\quad multi-answer                &  2{,}386 &  6.2\% \\
\midrule
Items with five options           & 38{,}279 & 100.0\% \\
Items with reference explanation  & 23{,}582 & 61.6\% \\
Items with source difficulty      & 38{,}217 & 99.8\% \\
\bottomrule
\end{tabular}
\end{table}

%
Each retained item is a tuple $(q, \mathcal{C}, y)$, where $q$ is the question, $\mathcal{C} = \{c_{\texttt{A}}, \dots, c_{\texttt{E}}\}$ is the option set, and $y \subseteq \Sigma$ is the ground-truth answer set over the option alphabet $\Sigma = \{\texttt{A}, \texttt{B}, \texttt{C}, \texttt{D}, \texttt{E}\}$.
Every item has exactly five options, so the random-guess floor is uniform across the benchmark.
We include two types of question: the majority have a single correct option (35{,}893 items, $93.8\%$), and the remainder have two or more (2{,}386 items, $6.2\%$).
The multi-answer items are there to raise the difficulty of the testbed.
Table~\ref{tab:corpus} summarizes the composition.

\subsection{Human Reference}
\label{sec:human}

%
An accuracy figure on a knowledge benchmark is hard to interpret without a reference point.
Whether $85\%$ on licensing material counts as competence or failure can be judged only against the people who are licensed to act on the same knowledge.
We built the human reference into the testbed and collected it under conditions that match the model evaluation.

%
We first built a web application to conduct the study.
It authenticates each practitioner, assigns a fixed private block of items, presents one item at a time, and records every interaction.
Items were drawn as a stratified 6{,}000-item subset of the corpus and partitioned into 40 blocks of 150.
Practitioner $u$ receives block $\lfloor (u-1)/3 \rfloor$, so each block goes to exactly three practitioners and each practitioner answers 150 items.
Several screenshots from our annotation interface are provided in Appendix~\ref{app:interface}.

%
Each item is presented on its own screen, and a single submission records the answer together with all annotations, then advances to the next unanswered item.
Beyond the answer, the practitioner also assigns up to three \emph{domain labels} from a ten-category taxonomy, rates \emph{difficulty} on a five-point scale, and may \emph{flag} the item as incorrect.
The difficulty question is subjective: how hard the item was \emph{for that rater}.

%
The ten-domain taxonomy (Table~\ref{tab:domains}) was defined in consultation with our practitioners, so that it partitions TCM practice the way the field is actually organized.
An item can belong to more than one domain: choosing a formula for a gynecological presentation, for example, involves both materia medica and internal medicine.
We therefore allow multiple labels, though $82.3\%$ of individual responses use exactly one.
The resulting mix follows the licensing curriculum: Internal Medicine is assigned to $41.3\%$ of subset items and Materia Medica to $31.1\%$, while ENT and Ophthalmology is assigned to only 84 items ($1.7\%$).

%
This study enrolled 101 licensed practitioners, who produced 15{,}151 responses over 5{,}250 distinct items.
Of these, 4{,}801 items received exactly three independent responses.
These 4{,}801 three-rater items form the \emph{comparison subset} on which all human--model analysis is performed; the rest are released but excluded from aggregate human accuracy.
We take a \emph{majority vote} over the three annotators, which reduces each item to a single correct/incorrect label and makes practitioner performance directly comparable to a model's score on the same items.
Practitioners spent a median of $23.5$ seconds per item, and their difficulty ratings span the full scale (see Table~\ref{tab:humandiff}).
The ratings also track performance: accuracy falls steadily from $74.7\%$ on items rated easiest to $40.4\%$ on items rated hardest.

\begin{table}[t]
\centering
\caption{
Practitioner-rated difficulty over all 15{,}151 responses.
\emph{Accuracy} is the share of responses at each rating that were correct, and \emph{Time} is the median time spent on the item before submission.
Ratings are per response, so an item answered by three practitioners contributes three ratings and may receive a different one from each.
}
\label{tab:humandiff}
\begin{tabular}{lrrrr}
\toprule
Rating & Responses & Share & Accuracy & Time \\
\midrule
1 (easiest) &  3{,}205 &  21.2\% & 74.7\% & 20.3\,s \\
2           &  3{,}611 &  23.8\% & 69.7\% & 24.5\,s \\
3           &  5{,}128 &  33.8\% & 59.2\% & 24.9\,s \\
4           &  1{,}914 &  12.6\% & 49.3\% & 24.0\,s \\
5 (hardest) &  1{,}293 &   8.5\% & 40.4\% & 23.6\,s \\
\midrule
All         & 15{,}151 & 100.0\% & 62.1\% & 23.5\,s \\
\bottomrule
\end{tabular}
\end{table}

\subsection{Evaluation Protocol}
\label{sec:harness}

%
We run our testbed on the \textsc{vLLM} inference engine, using greedy decoding at temperature zero with a 16-token budget.
From a single generation per item, the harness extracts two predictions.
The \emph{generative} prediction $\hat{y}$ is read from the decoded text: the harness scans it for characters in $\Sigma$, then deduplicates them and sorts them into a canonical set.
The \emph{log-probability} prediction is taken from the distribution over the first generated token,
\begin{equation}
    \hat{y}^{\,\mathrm{lp}} \;=\; \argmax_{c \,\in\, \Sigma} \; \ell(c),
    \qquad
    m \;=\; \ell_{(1)} - \ell_{(2)},
    \label{eq:lp}
\end{equation}
where $\ell(c)$ is the log-probability of option letter $c$, set to $-\infty$ for letters outside the returned top-$k$ set, and $m$ is the \emph{margin} between the top two letters.

%
The second approach separates knowledge from instruction-following.
A model that knows the answer but cannot produce the requested format is scored wrong on the generative channel and right on the log-probability one.
The gap between the two therefore reveals a formatting failure instead of letting it be mistaken for ignorance.
The margin gives us a confidence signal, and it is the basis for the calibration analysis of Appendix~\ref{app:calibration}.

%
Our primary metric is exact match (prediction accuracy), $\ind{\hat{y} = y}$.
We apply it the same way to single- and multi-answer items, so a multi-answer item counts as correct only when the model selects every correct option and no incorrect one.
For multi-answer items, we also report set-overlap metrics:
\begin{equation}
    \mathrm{P} = \frac{|\hat{y} \cap y|}{|\hat{y}|}, \qquad
    \mathrm{R} = \frac{|\hat{y} \cap y|}{|y|}, \qquad
    \mathrm{F}_1 = \frac{2\,\mathrm{P}\,\mathrm{R}}{\mathrm{P} + \mathrm{R}}.
    \label{eq:setmetrics}
\end{equation}
Precision and recall separate two ways of being wrong: a model that names too few options is conservative, and one that names too many is indiscriminate.
\section{Experiments}
\label{experiment}

%
We use \textsc{TCMQA} to evaluate 29 instruction-tuned models from 9 families under the protocol of Section~\ref{sec:harness}, asking how good current models are at TCM, how they compare against licensed practitioners, and which parts of the field they handle well.
Five findings follow.
\begin{itemize}[leftmargin=1.4em, itemsep=0.25em, topsep=0.35em]
    \item The benchmark spans a wide range of difficulty: it separates models across 59 accuracy points, from $18.95\%$ to $77.98\%$, and no model comes close to saturating it (Section~\ref{sec:overall}).
    \item Pretraining exposure to Chinese-language material matters far more than parameter count: the 12B Gemma is beaten by a Qwen model an eighth its size (Section~\ref{sec:scale}).
    \item Every model loses accuracy on multi-answer items, and exact match hides why: weak models answer as though a single option were wanted (Section~\ref{sec:format}).
    \item Nine of the 29 models outperform the practitioner majority vote of $64.94\%$ on the shared items, and all nine come from the Qwen family (Section~\ref{sec:human-compare}).
    \item Practitioners and models do not find the same items hard: model accuracy is flat across practitioner consensus and practitioner-rated difficulty (Section~\ref{sec:orthogonality}).
\end{itemize}

\begin{table}[!tp]
\centering
\caption{\textbf{Full leaderboard.} Models are grouped by family and ordered by size within each family. \emph{Full} is accuracy on all 38{,}279 items; \emph{Single} and \emph{Multi} split that by answer type. \emph{Subset} is accuracy on the 4{,}801 items practitioners also answered, and $\Delta$ is the difference from their majority vote of $64.94\%$, in bold where a model wins. All numbers are exact-match percentages.}
\label{tab:leaderboard}
\small
\definecolor{famrow}{gray}{0.90}
\begin{tabular}{lrrrrrr}
\toprule
Model & Params & Full & Single & Multi & Subset & $\Delta$ human \\
\midrule
\rowcolor{famrow}
\multicolumn{7}{l}{\bfseries Gemma2} \\
\quad Gemma-2-2B-It      &  2.61 & 25.09 & 26.44 &  4.78 & 26.74 & $-38.20$ \\
\quad Gemma-2-9B-It      &  9.24 & 36.93 & 38.72 & 10.10 & 37.78 & $-27.16$ \\
\addlinespace
\rowcolor{famrow}
\multicolumn{7}{l}{\bfseries Gemma3} \\
\quad Gemma-3-270M-It    &  0.27 & 18.95 & 20.20 &  0.00 & 17.93 & $-47.01$ \\
\quad Gemma-3-1B-It      &  1.00 & 23.82 & 25.17 &  3.39 & 21.85 & $-43.09$ \\
\quad Gemma-3-4B-It      &  4.30 & 30.19 & 31.54 &  9.97 & 30.31 & $-34.63$ \\
\quad Gemma-3-12B-It     & 12.19 & 39.60 & 41.03 & 18.11 & 40.93 & $-24.01$ \\
\addlinespace
\rowcolor{famrow}
\multicolumn{7}{l}{\bfseries Gemma4} \\
\quad Gemma-4-E2B-It     &  5.12 & 33.43 & 34.68 & 14.54 & 36.80 & $-28.14$ \\
\quad Gemma-4-E4B-It     &  8.00 & 44.69 & 46.65 & 15.09 & 46.43 & $-18.51$ \\
\addlinespace
\rowcolor{famrow}
\multicolumn{7}{l}{\bfseries OLMo2} \\
\quad OLMo-2-1B-Inst.    &  1.48 & 21.05 & 22.45 &  0.00 & 20.83 & $-44.11$ \\
\quad OLMo-2-7B-Inst.    &  7.30 & 24.36 & 25.72 &  3.86 & 25.83 & $-39.11$ \\
\quad OLMo-2-13B-Inst.   & 13.72 & 26.33 & 27.27 & 12.11 & 26.37 & $-38.57$ \\
\addlinespace
\rowcolor{famrow}
\multicolumn{7}{l}{\bfseries OLMo3} \\
\quad OLMo-3-7B-Inst.    &  7.30 & 26.82 & 28.18 &  6.33 & 26.85 & $-38.09$ \\
\addlinespace
\rowcolor{famrow}
\multicolumn{7}{l}{\bfseries Qwen1.5} \\
\quad Qwen1.5-0.5B-Chat  &  0.62 & 23.99 & 25.33 &  3.90 & 23.89 & $-41.05$ \\
\quad Qwen1.5-1.8B-Chat  &  1.84 & 38.53 & 40.63 &  6.96 & 44.37 & $-20.57$ \\
\quad Qwen1.5-4B-Chat    &  3.95 & 49.85 & 52.31 & 12.82 & 56.61 & $-8.33$  \\
\quad Qwen1.5-7B-Chat    &  7.72 & 61.23 & 63.78 & 22.88 & 70.71 & $\mathbf{+5.77}$  \\
\quad Qwen1.5-14B-Chat   & 14.17 & 67.89 & 69.97 & 36.50 & 78.11 & $\mathbf{+13.17}$ \\
\addlinespace
\rowcolor{famrow}
\multicolumn{7}{l}{\bfseries Qwen2} \\
\quad Qwen2-0.5B-Inst.   &  0.49 & 35.15 & 36.49 & 14.96 & 39.49 & $-25.45$ \\
\quad Qwen2-1.5B-Inst.   &  1.54 & 60.81 & 63.70 & 17.23 & 71.88 & $\mathbf{+6.94}$  \\
\quad Qwen2-7B-Inst.     &  7.62 & 75.02 & 76.66 & 50.29 & 86.73 & $\mathbf{+21.79}$ \\
\addlinespace
\rowcolor{famrow}
\multicolumn{7}{l}{\bfseries Qwen2.5} \\
\quad Qwen2.5-0.5B-Inst. &  0.49 & 29.77 & 31.43 &  4.74 & 29.70 & $-35.24$ \\
\quad Qwen2.5-1.5B-Inst. &  1.54 & 52.31 & 54.28 & 22.55 & 57.78 & $-7.16$  \\
\quad Qwen2.5-3B-Inst.   &  3.09 & 59.27 & 61.78 & 21.54 & 67.32 & $\mathbf{+2.38}$  \\
\quad Qwen2.5-7B-Inst.   &  7.62 & 71.59 & 73.58 & 41.66 & 81.48 & $\mathbf{+16.54}$ \\
\quad Qwen2.5-14B-Inst.  & 14.77 & 77.98 & 79.51 & 54.99 & 86.38 & $\mathbf{+21.44}$ \\
\addlinespace
\rowcolor{famrow}
\multicolumn{7}{l}{\bfseries Qwen3.5} \\
\quad Qwen3.5-0.8B       &  0.87 & 36.82 & 38.27 & 14.92 & 41.12 & $-23.82$ \\
\quad Qwen3.5-2B         &  2.27 & 50.77 & 53.82 &  4.90 & 56.51 & $-8.43$  \\
\quad Qwen3.5-4B         &  4.66 & 67.37 & 69.59 & 34.03 & 73.57 & $\mathbf{+8.63}$  \\
\quad Qwen3.5-9B         &  9.65 & 75.02 & 77.06 & 44.30 & 81.44 & $\mathbf{+16.50}$ \\
\bottomrule
\end{tabular}
\end{table}

\subsection{Overall Performance}
\label{sec:overall}

\paragraph{Setup.}
We first ask whether our benchmark separates models at all.
We evaluate 29 instruction-tuned models from 9 families, ranging from 0.27B to 14.77B parameters, under the protocol of Section~\ref{sec:harness}.
Most are small or mid-sized open-weight models, the kind that ordinary users can run locally, which matters in the medical domain, where privacy concerns push users toward local models.
We include both Chinese- and Western-developed models.
All 29 are scored by exact match on the full corpus of 38{,}279 items, reported overall and split by answer type.

\begin{figure}[tbp]
\centering
\includegraphics[width=\linewidth]{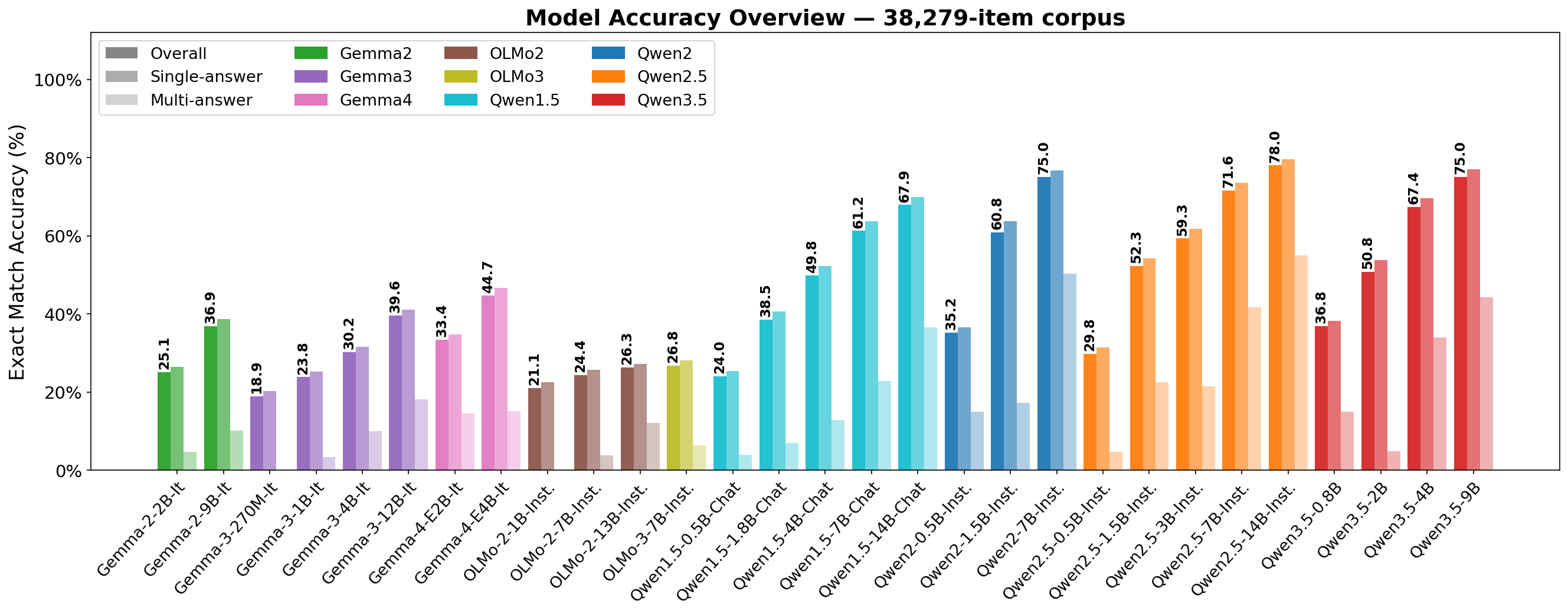}
\caption{\textbf{Model accuracy by answer type.} Overall, single-answer, and multi-answer accuracy for all 29 models, grouped by family. The multi-answer bar is shorter than the single-answer bar for all 29 models, and the gap grows as models get weaker, so the overall score hides this weakness.}
\label{fig:overview}
\end{figure}

\paragraph{Results.}
Table~\ref{tab:leaderboard} and Figure~\ref{fig:overview} report accuracy for all 29 models.
Two findings stand out.
First, every model does worse on multi-answer questions than on single-answer ones, and the gap widens as models get weaker.
Second, the benchmark spreads models out: scores range from $18.95\%$ to $77.98\%$.
The best model, Qwen2.5-14B-Inst., gets about four questions in five right, while the worst, Gemma-3-270M-It, performs no better than random guessing.
Two further patterns preview the sections that follow: the Qwen models (Qwen3.5, Qwen2.5, Qwen2, and even Qwen1.5) sit at the frontier of the leaderboard, ahead of Western-developed families such as Gemma and OLMo, and nine of the 29 models outperform the practitioner reference---all nine of them Qwen.

\begin{figure}[tbp]
\centering
\includegraphics[width=0.8\linewidth]{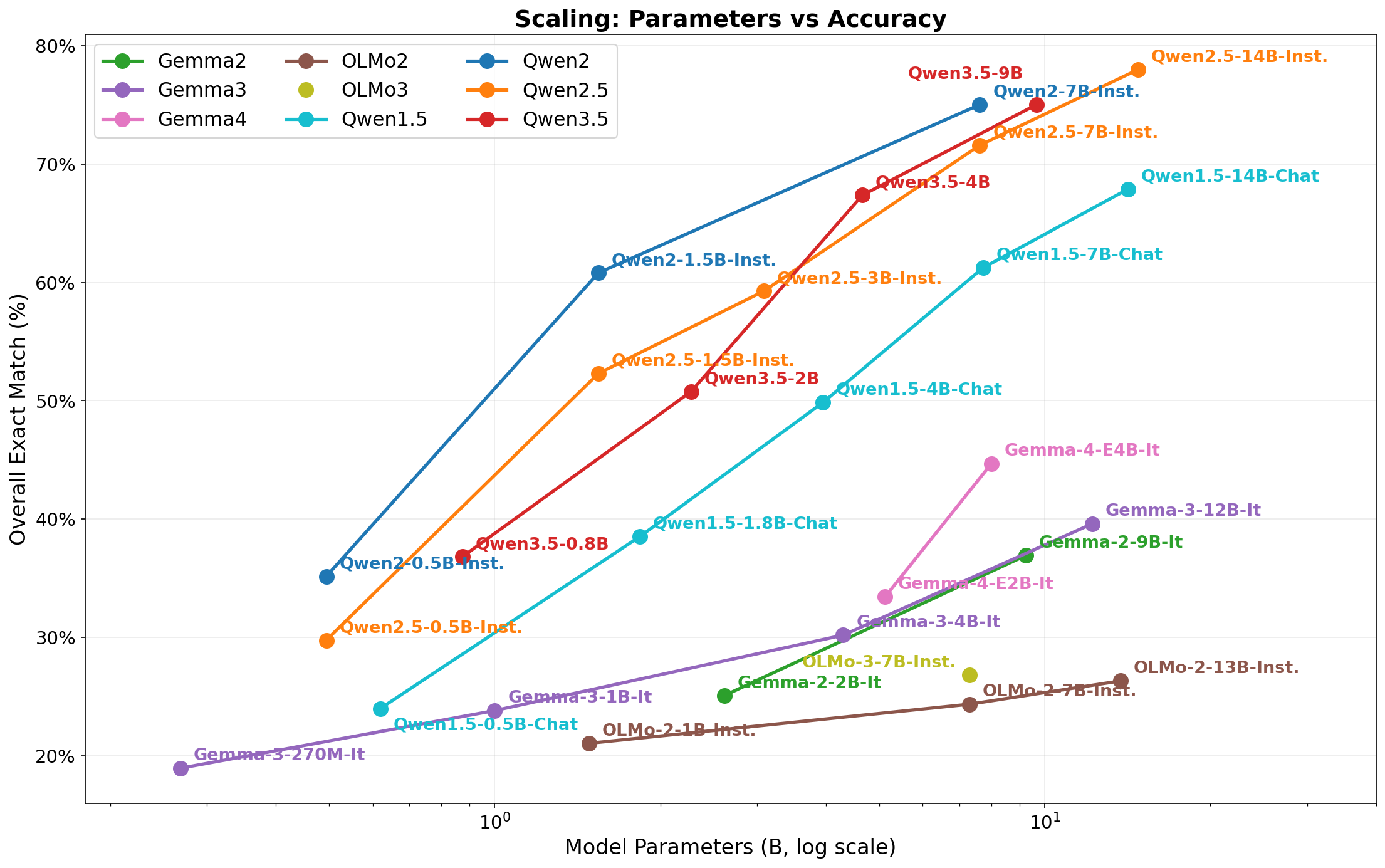}
\caption{\textbf{Accuracy against model size.} Accuracy against parameter count on a log axis, one line per family. The Qwen lines sit in a band of their own, and the gap to Gemma and OLMo grows with size instead of closing. Note also that Qwen2 beats Qwen2.5 at $0.5$B, $1.5$B, and $7$B; Qwen2.5 wins only at $14$B, where Qwen2 has no model, so Qwen2.5's top spot comes from being bigger rather than better.}
\label{fig:scaling}
\end{figure}

\subsection{Training Data Matters More Than Size}
\label{sec:scale}

\paragraph{Setup.}
Section~\ref{sec:overall} established that the benchmark separates models across 59 accuracy points.
We now ask what explains that spread.
We plot accuracy against parameter count for all 29 models on a logarithmic axis, one line per family (Figure~\ref{fig:scaling}).
The families span 0.27B to 14.77B parameters, and six of the nine cover at least three sizes, so both the within-family and the across-family comparison are available over a usable range.

\paragraph{Results.}
First, within every family, accuracy rises monotonically with size: once the training data is held fixed, scale buys real gains.
Across families, however, scale stops predicting accuracy.
Gemma-3-12B-It reaches $39.60\%$, while Qwen2-1.5B-Inst., eight times smaller, reaches $60.81\%$; with the single exception of Qwen1.5-1.8B-Chat, every Qwen model at 1.5B or above beats the 12B Gemma.
The OLMo family never passes $27\%$ at any size we tried, and growing from 1B to 13B gains it only about five points.
Every model that outperforms the practitioner baseline is a Qwen model.
Second, newer generations narrow the gap without closing it.
Qwen3.5 climbs steadily from $36.82\%$ to $75.02\%$ across our four sizes, and Qwen3.5-4B matches Qwen1.5-14B-Chat with a third of the parameters.
Gemma-4-E4B-It is the strongest Gemma we measured, and it still loses to Qwen3.5-2B, a model $3.5\times$ smaller from a family two generations older.
Third, progress is not uniformly forward, either: Qwen2.5 scores below Qwen2 at all three sizes where both exist.
The most plausible explanation is the composition of the pretraining corpus, since the families that do well are the ones trained on far more Chinese text.

\subsection{Answer Sets in Multi-Answer Items}
\label{sec:format}

\paragraph{Setup.}
Section~\ref{sec:overall} showed that every model loses accuracy on multi-answer items.
We look more closely at those items here.
We score the 2{,}386 multi-answer items with precision, recall, and F$_1$ alongside exact match, and record $|\hat{y}|$, the mean number of options a model names, against a mean answer size of $3.50$.

\begin{figure}[tbp]
\centering
\includegraphics[width=\linewidth]{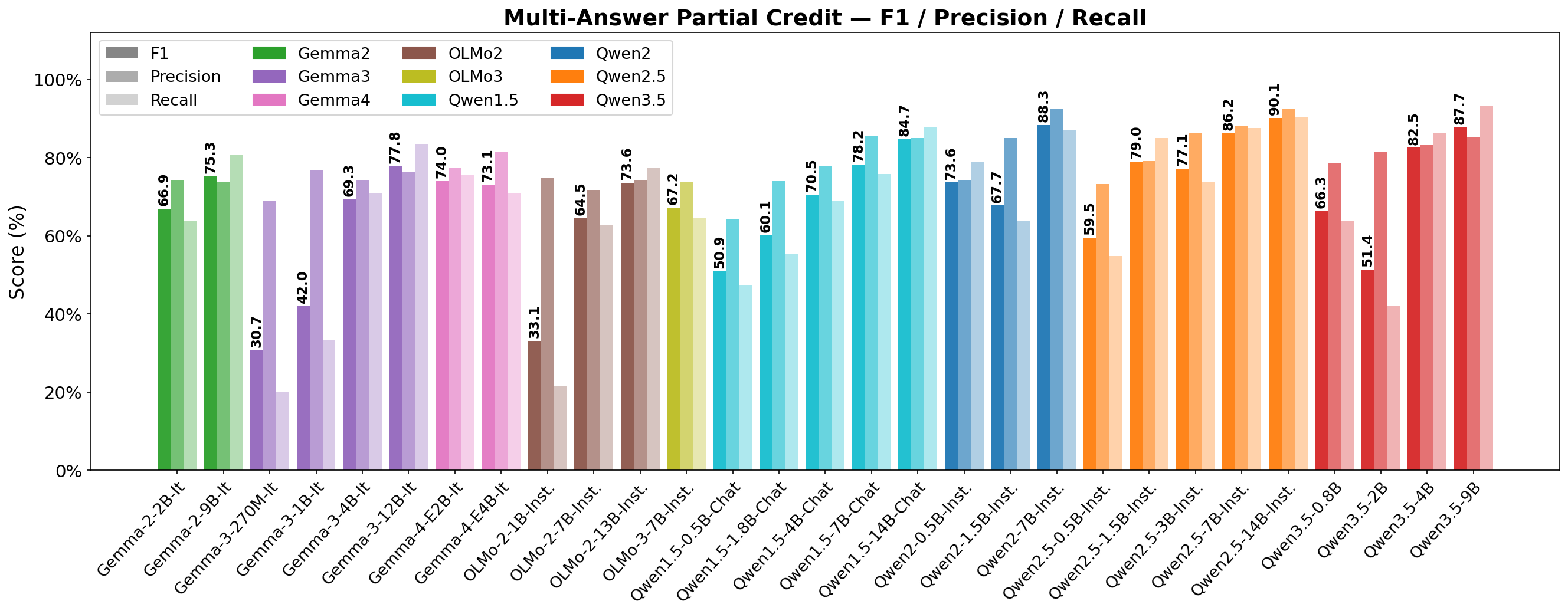}
\caption{\textbf{Precision, recall and F$_1$ on multi-answer items.} Set-overlap metrics on the 2{,}386 multi-answer items. Precision beats recall for most models. That means they name too few options and treat these questions as single-answer.}
\label{fig:multi}
\end{figure}

\paragraph{Results.}
Multi-answer items hurt far more than their $6.2\%$ share suggests (Table~\ref{tab:multi}).
The best model falls from $79.51\%$ on single-answer items to $54.99\%$ on multi-answer ones, and models below $55\%$ overall often land in single digits.
The failures are systematic, and they run in two directions (Figure~\ref{fig:multi}).
Strong models answer about the right number of options: Qwen2.5-14B-Inst.\ answers $3.40$ with precision and recall both above $90\%$, so it picks a slightly wrong set rather than a set of the wrong size.
Weak models name too few options or too many.
Qwen3.5-2B names $1.84$, buying a precision of $81.3\%$ at a recall of $42.2\%$; Gemma-3-270M-It names $0.90$, as though a single option were wanted; and Gemma-3-12B-It overshoots at $3.82$.
Exact match scores these habits identically, so it understates what a model that names too few options actually knows.

\begin{table}[tbp]
\centering
\caption{\textbf{Multi-answer decomposition.} Results on the 2{,}386 multi-answer items. $|\hat{y}|$ is the mean number of options predicted; the answers contain $3.50$ on average. Precision above recall means the model named too few options, recall above precision means too many.}
\label{tab:multi}
\small
\begin{tabular}{lrrrrrr}
\toprule
Model & Single EM & Multi EM & P & R & F$_1$ & $|\hat{y}|$ \\
\midrule
Qwen2.5-14B-Inst. & 79.5 & 55.0 & 92.4 & 90.4 & 90.1 & 3.40 \\
Qwen2-7B-Inst.    & 76.7 & 50.3 & 92.5 & 86.9 & 88.3 & 3.25 \\
Qwen3.5-9B        & 77.1 & 44.3 & 85.3 & 93.2 & 87.7 & 3.84 \\
Qwen3.5-2B        & 53.8 &  4.9 & 81.3 & 42.2 & 51.4 & 1.84 \\
Gemma-3-12B-It    & 41.0 & 18.1 & 76.3 & 83.4 & 77.8 & 3.82 \\
OLMo-2-13B-Inst.  & 27.3 & 12.1 & 74.2 & 77.2 & 73.6 & 3.61 \\
Gemma-3-270M-It   & 20.2 &  0.0 & 69.0 & 20.2 & 30.7 & 0.90 \\
\bottomrule
\end{tabular}
\end{table}

\subsection{Comparing Against Licensed Practitioners}
\label{sec:human-compare}

\paragraph{Setup.}
We now turn to the human reference, and ask how open-weight models compare against the people licensed to practice.
We restrict to the 4{,}801 items that three practitioners each answered and reduce the three responses to a majority vote, which yields a single correct-or-incorrect label per item at $64.94\%$ accuracy.
All 29 models are scored on those same items, so both sides see identical questions, which is what makes the comparison fair even though the subset is smaller than the full corpus.

\begin{figure}[tbp]
\centering
\includegraphics[width=\linewidth]{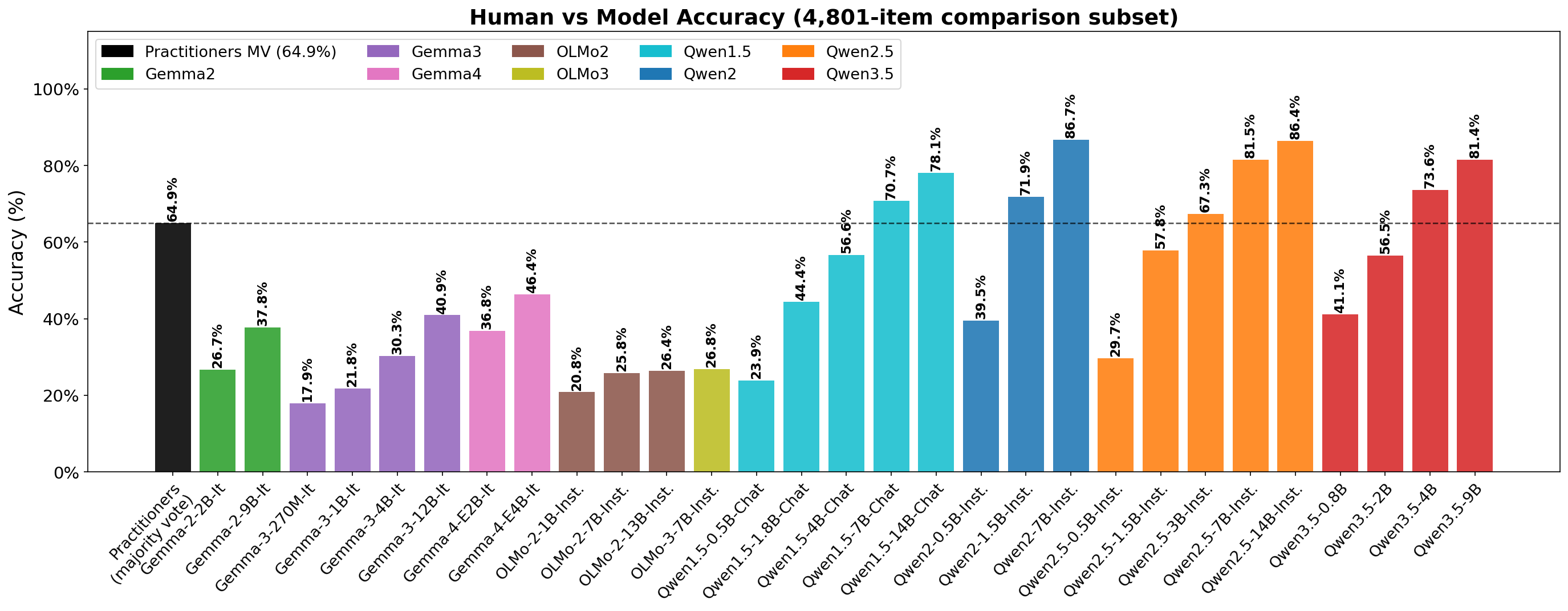}
\caption{\textbf{Models against the practitioner reference.} Accuracy on the 4{,}801 shared items: the practitioner majority vote against all 29 models, with the practitioner score as a dashed line. Nine models clear the line and all nine are Qwen. Every Gemma and OLMo falls short, including models larger than the Qwen ones that succeed.}
\label{fig:humanvsllm}
\end{figure}

\paragraph{Results.}
Nine of 29 models beat the practitioner majority vote (Figure~\ref{fig:humanvsllm}).
Qwen2-7B-Inst.\ leads at $86.73\%$, ahead by $21.79$ points, and the margin stays positive down to Qwen2.5-3B-Inst.
All nine are Qwen models, and the twenty below the line include every Gemma and every OLMo we tested, several of them larger than the Qwen models above the line.

\begin{figure}[tbp]
\centering
\includegraphics[width=0.75\linewidth]{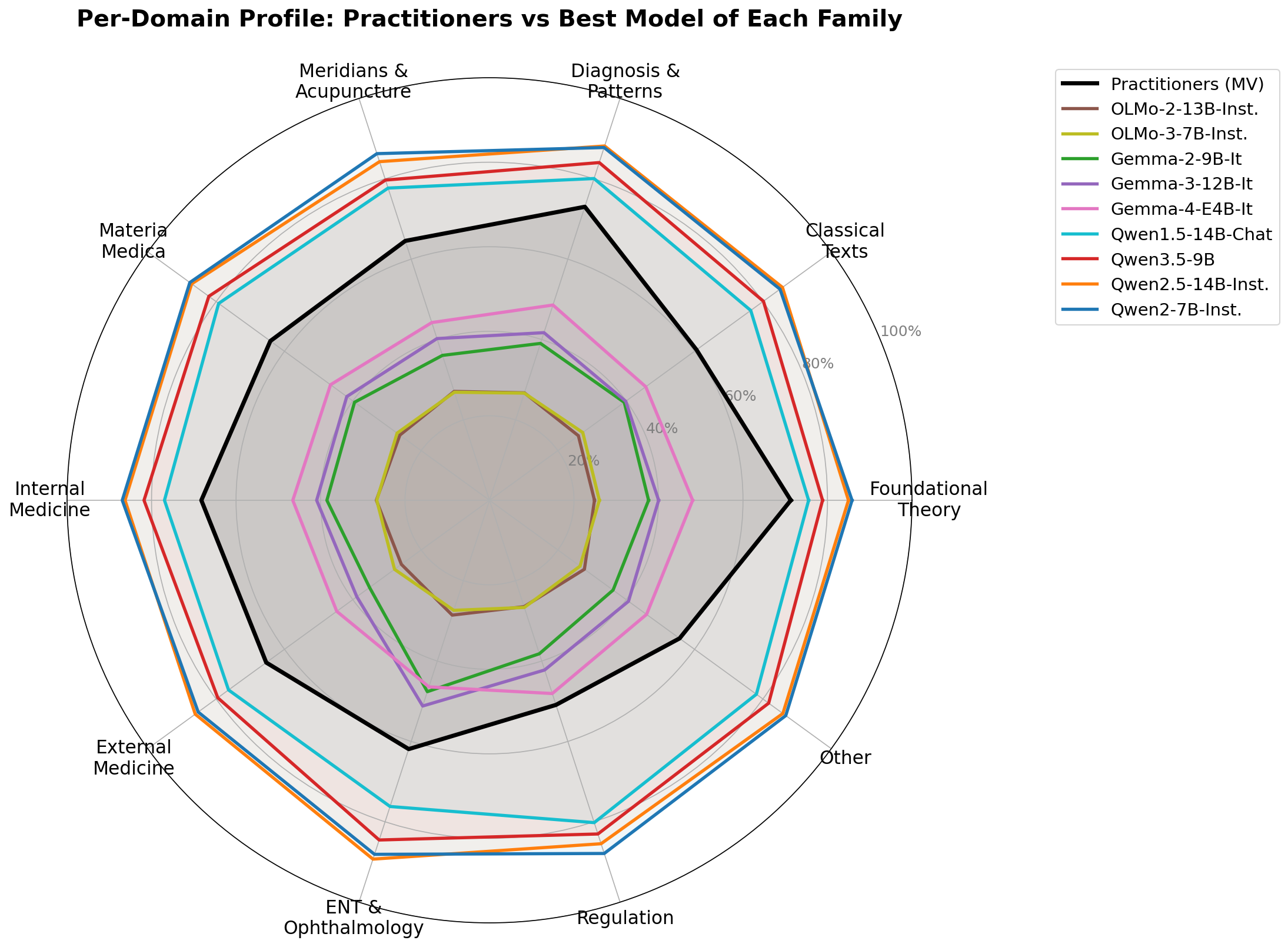}
\caption{\textbf{Per-domain accuracy.} Accuracy across the ten practitioner-assigned domains: the practitioner majority vote against the best model of each family. The order is the same on every axis. The three Qwen contours sit outside the practitioner contour everywhere, and every Gemma and OLMo contour sits inside it. The model contours are nearly circular while the practitioner contour is lopsided, so the gap is shaped by the human side.}
\label{fig:radar}
\end{figure}

%
The advantage does not come from one weak area (Figure~\ref{fig:radar}).
The three best models beat the practitioners in all ten domains.
What changes from domain to domain is the size of the gap, and it changes because practitioner scores do.
Practitioners are weakest on Regulation at $50.9\%$, where the best model reaches $87.9\%$, and strongest on Diagnosis and Patterns at $73.0\%$ and Foundational Theory at $71.3\%$, where the same model is only about 15 points ahead.
Model accuracy hardly moves across domains, so the models are furthest ahead on material that is mostly memorization and closest on material that needs clinical reasoning.

\subsection{Models and Practitioners Find Different Things Hard}
\label{sec:orthogonality}

\paragraph{Setup.}
Nine models outscore the practitioners, but a higher score need not mean the same ability.
We therefore ask whether the two sides (models and humans) find the same items hard.
On the shared subset we relate model accuracy to two practitioner signals, how many of the three answered each item correctly and the difficulty they rated it before learning the outcome, and we measure item-level agreement with Cohen's $\kappa$.
Flat accuracy against the first two signals would mean the two kinds of difficulty are unrelated, but only if the practitioner signal carries information at all.
We test that first against $\mathrm{Binomial}(3, 0.6248)$, the distribution of correct counts had the three practitioners answered independently at their observed rate.

\paragraph{Results.}
The practitioner signal is real (Table~\ref{tab:consensus}).
Items that all three practitioners miss are $2.43$ times more common than independence predicts and items all three answer are $1.45$ times more common, giving $\chi^2 = 1009.0$ on three degrees of freedom against a critical value of $16.27$.
Practitioners agree about which questions are hard, far more than chance would allow.

\begin{table}[tbp]
\centering
\caption{\textbf{Practitioner agreement against independence.} Agreement is far stronger than independent answering would produce. Expected counts come from $\mathrm{Binomial}(3, 0.6248)$ over the 4{,}801 items with three raters.}
\label{tab:consensus}
\small
\begin{tabular}{lrrr}
\toprule
Practitioners correct & Observed & If independent & Ratio \\
\midrule
0 of 3 &   616 &  253.6 & $2.43\times$ \\
1 of 3 & 1{,}067 & 1{,}266.8 & $0.84\times$ \\
2 of 3 & 1{,}422 & 2{,}109.6 & $0.67\times$ \\
3 of 3 & 1{,}696 & 1{,}171.0 & $1.45\times$ \\
\midrule
\multicolumn{4}{l}{$\chi^2 = 1009.0$ (3 df), Fleiss' $\kappa = 0.263$} \\
\bottomrule
\end{tabular}
\end{table}

\begin{figure}[tbp]
\centering
\includegraphics[width=\linewidth]{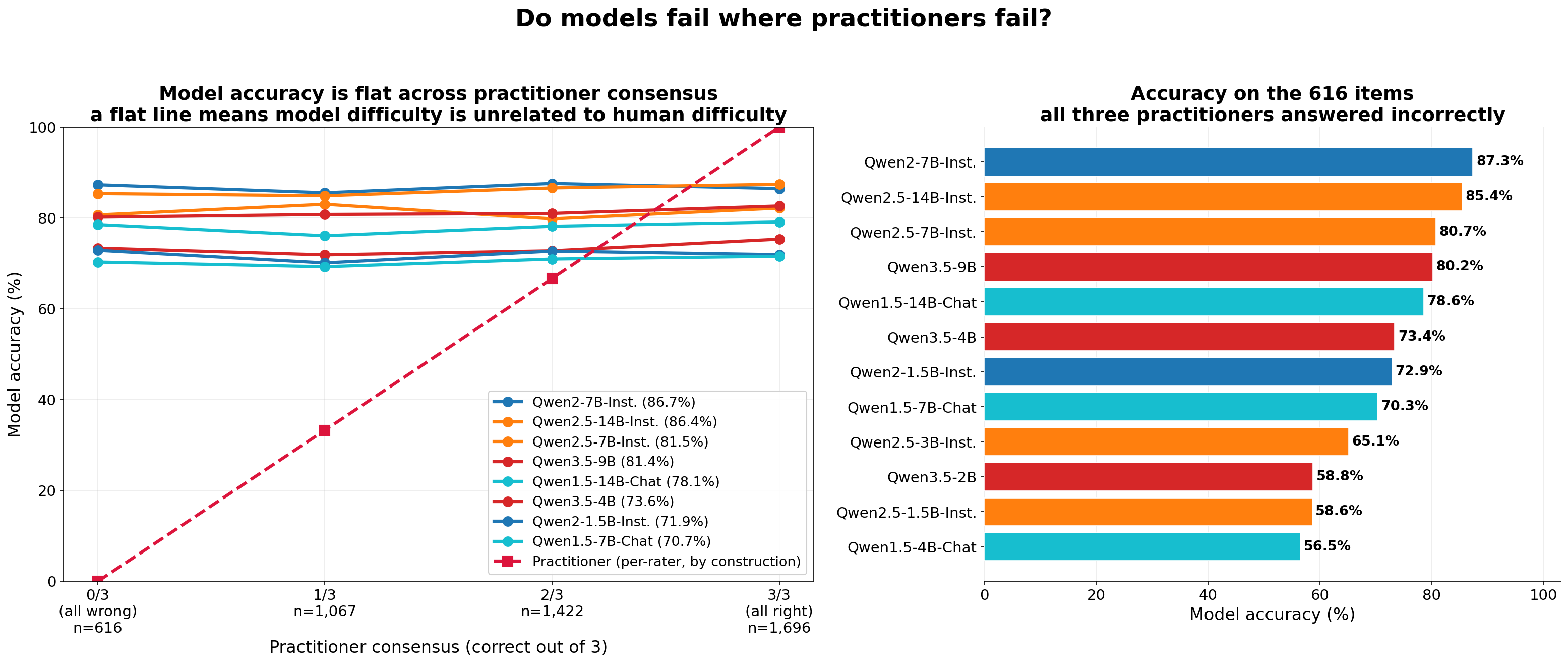}
\caption{\textbf{Model accuracy by practitioner consensus.} Left: model accuracy by how many of the three practitioners answered correctly. Right: accuracy on the 616 items all three got wrong. The model lines are flat, and the best model does slightly better on the questions every practitioner missed; the dashed practitioner line runs from $0\%$ to $100\%$ by construction. Because practitioner agreement is real rather than noise (Table~\ref{tab:consensus}), flat lines mean the two difficulties are unrelated.}
\label{fig:consensus}
\end{figure}

\begin{figure}[tbp]
\centering
\includegraphics[width=0.7\linewidth]{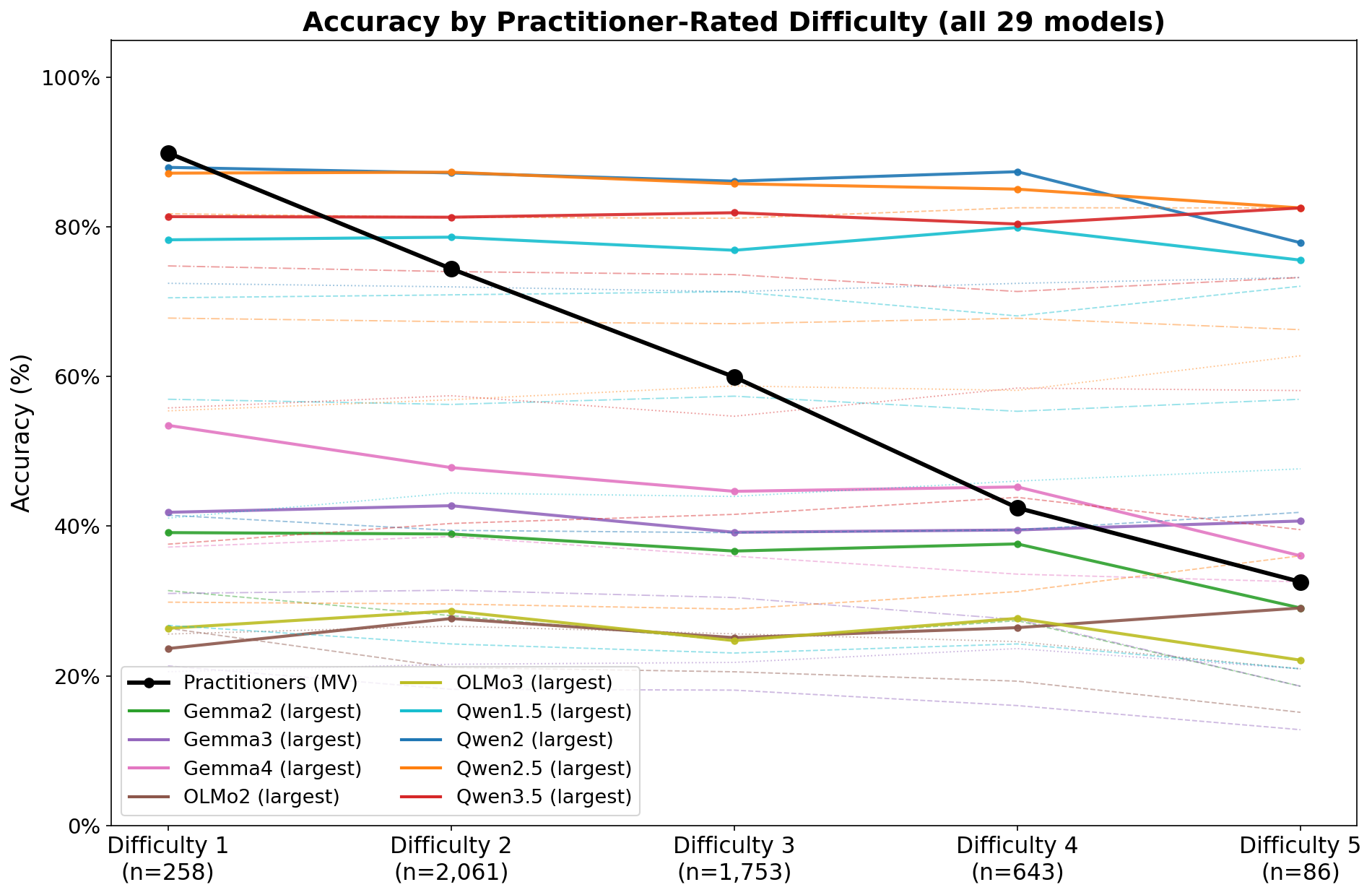}
\caption{\textbf{Accuracy across the difficulty scale.} All 29 models plus the practitioner line, over the practitioners' five-point scale. Practitioner accuracy falls 57 points while almost every model line stays flat: at level 1 the practitioners beat all but two models, and by level 5 they are below Gemma-4-E4B-It. The ratings were given before anyone knew the outcome (Section~\ref{sec:human}), so this is separate evidence rather than a restatement of Figure~\ref{fig:consensus}. Level 5 holds only 86 items, so read the last point loosely.}
\label{fig:difficulty}
\end{figure}

%
Model accuracy ought to follow that signal, and it does not (Figure~\ref{fig:consensus}).
On the 616 items every practitioner missed, Qwen2-7B-Inst.\ scores $87.34\%$, slightly \emph{above} its $86.73\%$ subset average; on the items every practitioner answered it scores $86.50\%$.
The same flatness holds for all 29 models, down to those near $20\%$.
The difficulty ratings agree (Figure~\ref{fig:difficulty}): practitioner accuracy falls 57 points, from $89.9\%$ at level 1 to $32.6\%$ at level 5, while over the same range Qwen2-7B-Inst.\ loses ten points, Qwen3.5-9B under two, and Gemma-3-12B-It stays between $39.2\%$ and $42.7\%$ throughout.
Gemma-4-E4B-It is the one real exception, and its $17.5$-point decline is still a third of the practitioners'.
Models overtake practitioners on hard items because the practitioners fall away, not because the models hold up.

\begin{figure}[tbp]
\centering
\includegraphics[width=0.8\linewidth]{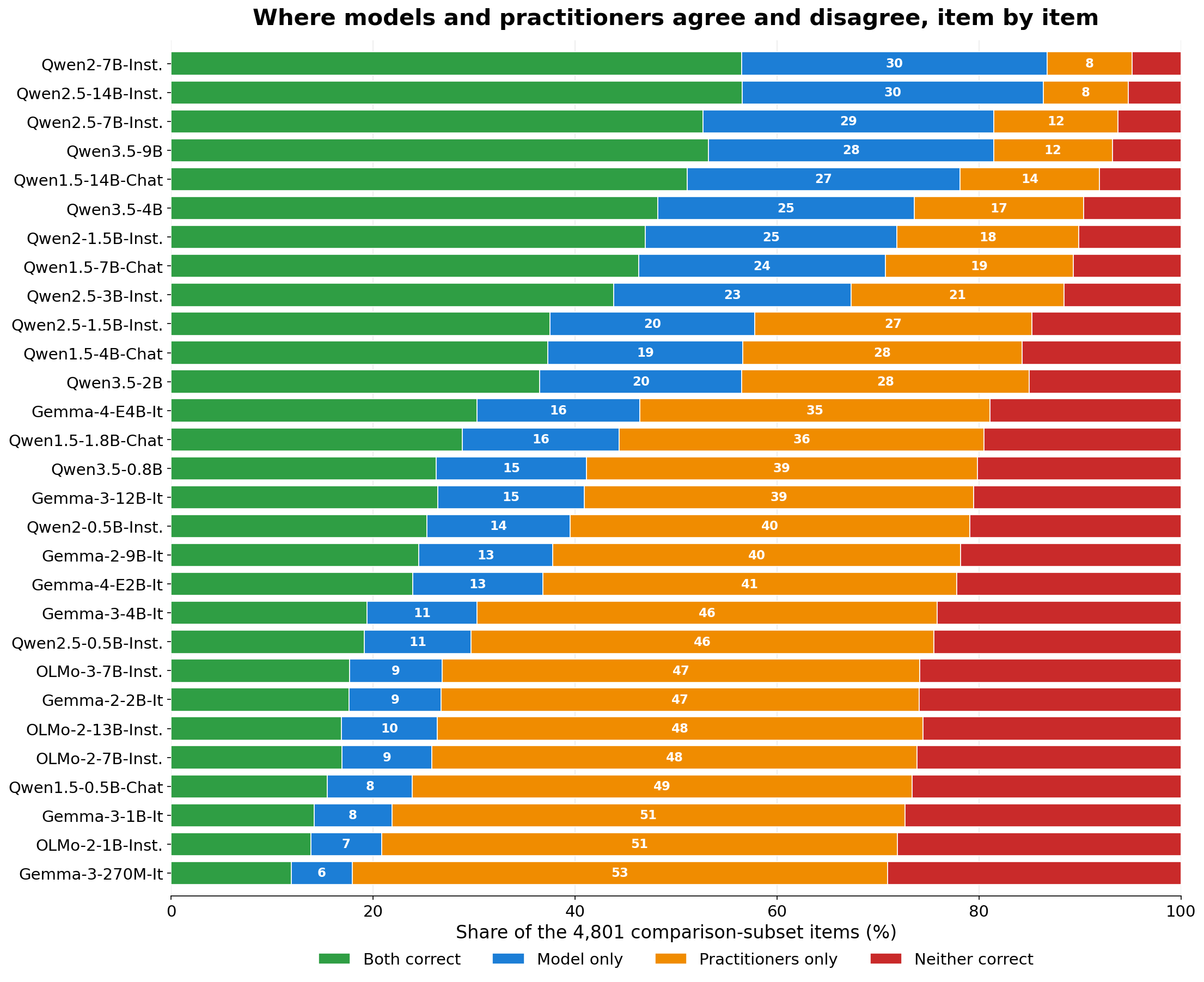}
\caption{\textbf{Item-by-item outcomes.} Outcomes against the practitioner majority vote for all 29 models, ordered by accuracy; the numbers inside the bars are the two disagreement cells as a share of the 4{,}801 items. Going from the worst model to the best raises model-only wins from $6\%$ to $30\%$, but practitioner-only wins fall from $53\%$ to $8\%$ and stop there. The orange band never closes, so a single ranked score hides more than it shows.}
\label{fig:contingency}
\end{figure}

%
Item-level agreement points the same way.
Cohen's $\kappa$ between a model and the practitioner majority runs from $-0.017$ to $+0.023$ across all 29 models, indistinguishable from zero and with no trend in model quality, while between pairs of models it runs from $0.06$ to $0.23$, up to ten times higher.
Models err like each other rather than like the practitioners, which points to habits picked up in pretraining rather than to intrinsically hard items.
Figure~\ref{fig:contingency} shows what that costs.
For Qwen2-7B-Inst., both sides answer $56.5\%$ of the items and neither answers $4.8\%$, but the model alone answers $30.2\%$ and the practitioners alone $8.4\%$.
If a model were simply a better practitioner, that last cell would shrink toward zero as accuracy rose.
It does not: across the leaderboard, practitioner-only wins fall from $53\%$ to $8\%$ and then stop, so even our best model leaves several hundred items that the practitioners answer and it misses.

\section{Conclusion}
\label{sec:conclusion}

%
\textsc{TCMQA} pairs 38{,}279 questions from Chinese TCM licensing examinations with 15{,}151 responses from 101 licensed practitioners.
Across 29 instruction-tuned models from 9 families, the benchmark separates the field by 59 accuracy points and no model comes close to saturating it.
What a model was pretrained on predicts TCM ability far better than how large it is.
Nine models clear the practitioner majority vote of $64.94\%$, the best by $21.79$ points, and every one of them comes from a single Chinese-pretrained family; Western-pretrained models several times larger fall more than twenty points short.

%
Practitioners agree about which items are hard, yet model accuracy is flat across their consensus and nearly flat across their difficulty ratings.
Agreement with the practitioner majority is indistinguishable from zero for all 29 models, while models agree with each other up to ten times more often, and even the leading model misses several hundred items the practitioners answer.
The models are furthest ahead on Regulation, where the material is largely memorization, and closest on Diagnosis and Patterns.

\newpage

\bibliography{reference}
\bibliographystyle{achemso}

\appendix

\section{Annotation Interface}
\label{app:interface}

\begin{figure}[htbp]
\centering
\includegraphics[width=0.5\linewidth]{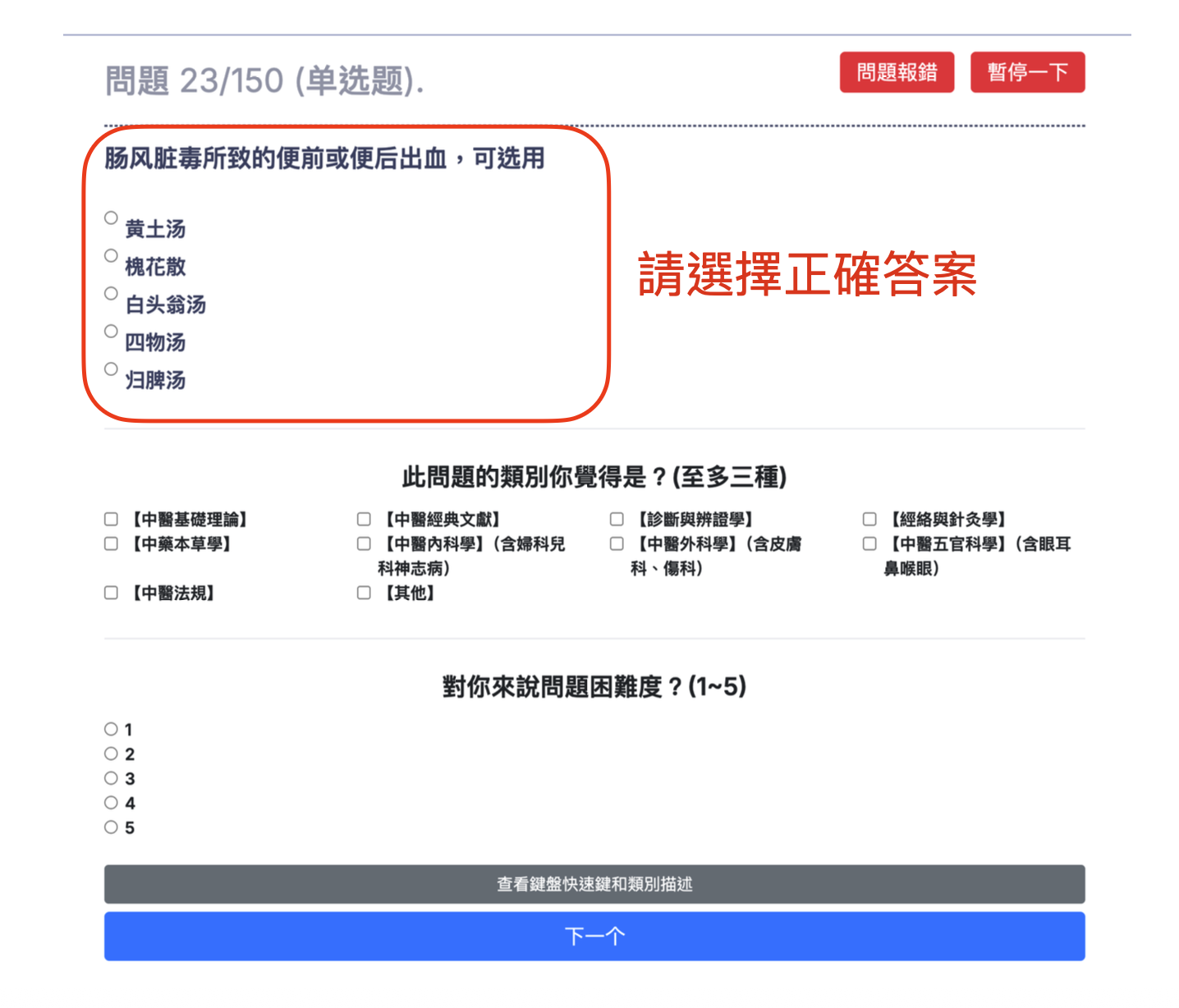}
\caption{\textbf{Answering an item.} The item is presented in its original Chinese with its option list, one item per screen, and the header shows position within the practitioner's 150-item block and whether the item is single- or multi-answer. Models see the same item text under the protocol of Section~\ref{sec:harness}, so both sides are answering identical questions.}
\label{fig:ui-answer}
\end{figure}

\begin{figure}[htbp]
\centering
\includegraphics[width=0.5\linewidth]{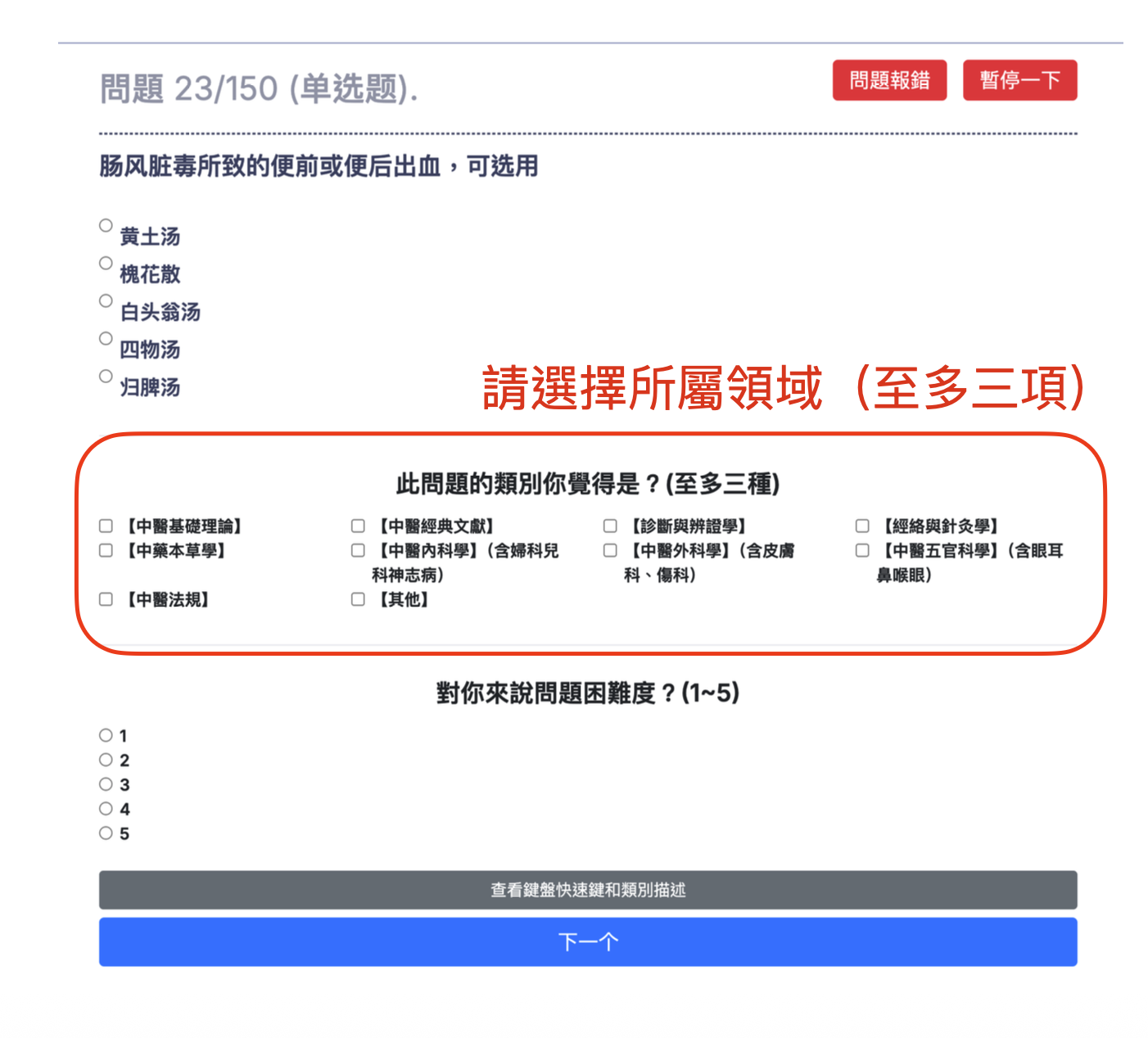}
\caption{\textbf{Assigning domain labels.} The practitioner tags the item with up to three of the ten domains of Table~\ref{tab:domains}, plus \zh{其他} (Other) for items the taxonomy does not cover. Multiple labels are allowed because one item can span domains, though $82.3\%$ of responses use exactly one. These labels produce the per-domain comparison of Figure~\ref{fig:radar}.}
\label{fig:ui-domain}
\end{figure}

\begin{figure}[htbp]
\centering
\includegraphics[width=0.5\linewidth]{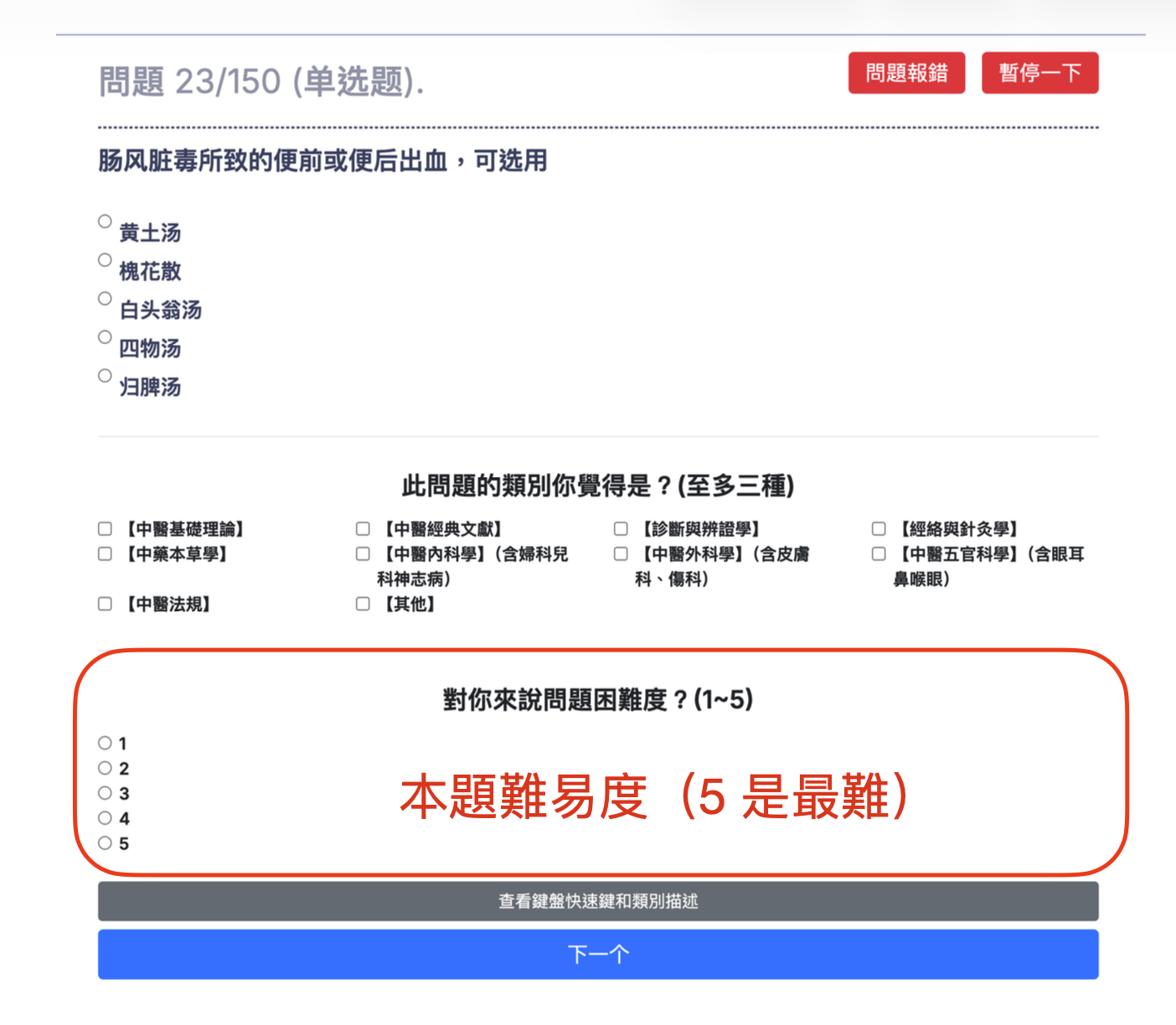}
\caption{\textbf{Rating difficulty.} A five-point scale, 5 being hardest. The question asks how hard the item was \emph{for that rater} rather than how hard it is in general, so the ratings are subjective by design. They are collected on the same screen as the answer and before any outcome is revealed, which is what lets Figure~\ref{fig:difficulty} treat them as independent evidence.}
\label{fig:ui-difficulty}
\end{figure}

\begin{figure}[htbp]
\centering
\includegraphics[width=0.5\linewidth]{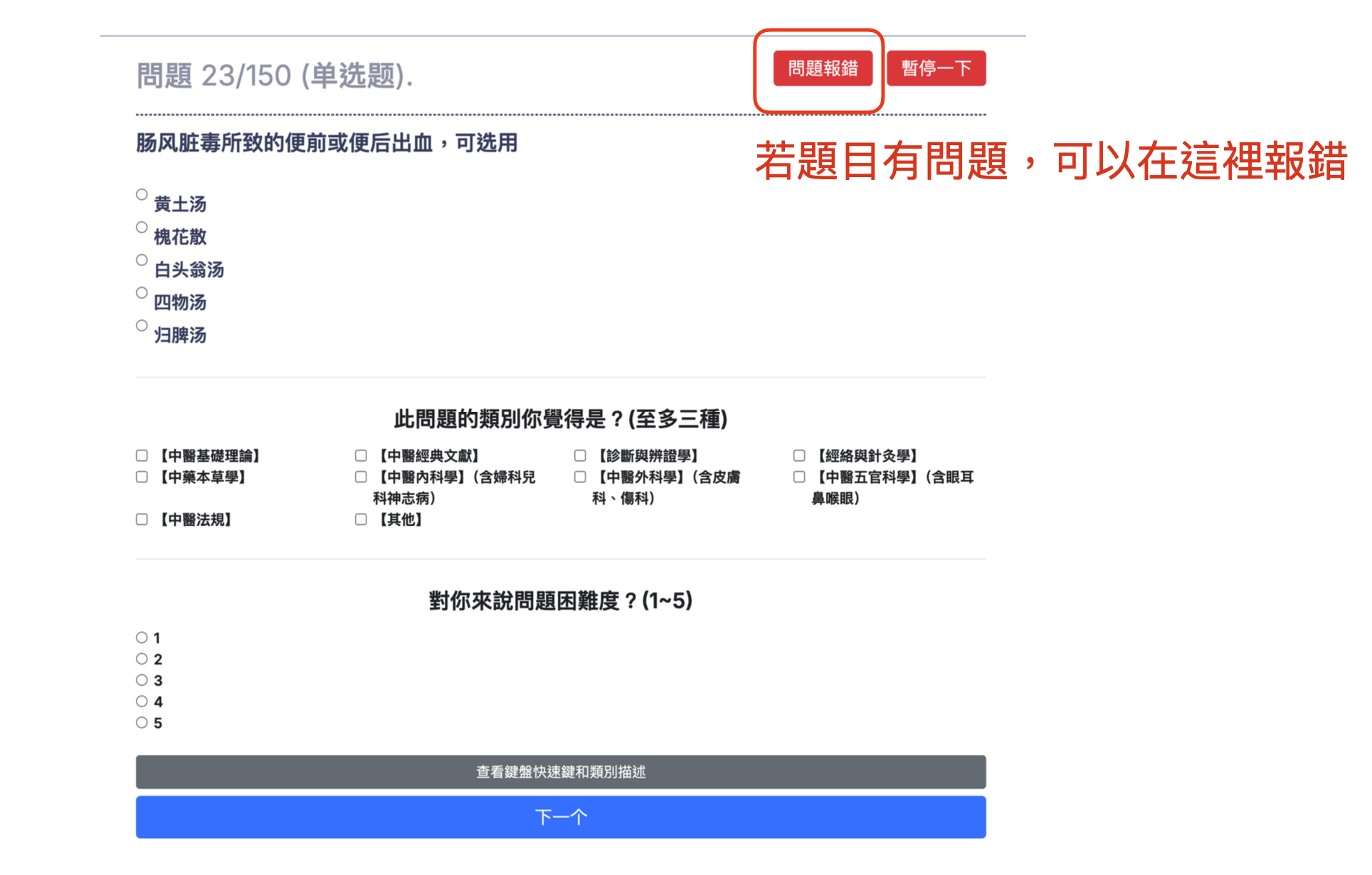}
\caption{\textbf{Flagging and pausing.} \zh{問題報錯} flags an item the practitioner believes is defective, and \zh{暫停一下} pauses the block. The flag gives a practitioner somewhere to report a bad item instead of guessing at it, and the pause keeps an interruption from being recorded as a slow or careless response; timing is measured per item, and the median is $23.5$ seconds.}
\label{fig:ui-flag}
\end{figure}

The human reference of Section~\ref{sec:human} was collected through a web application built for this study.
A practitioner signs in, is assigned a private block of 150 items, and sees one item per screen.
The screen carries four controls, shown one at a time in Figures~\ref{fig:ui-answer}--\ref{fig:ui-flag}, and a single submission records all of them together before advancing to the next unanswered item.

\FloatBarrier

\section{Model Confidence and Calibration}
\label{app:calibration}

Section~\ref{experiment} ranks models by how often they are right.
Here we ask a second question: when a model is wrong, does it know?

\begin{figure}[htbp]
\centering
\includegraphics[width=\linewidth]{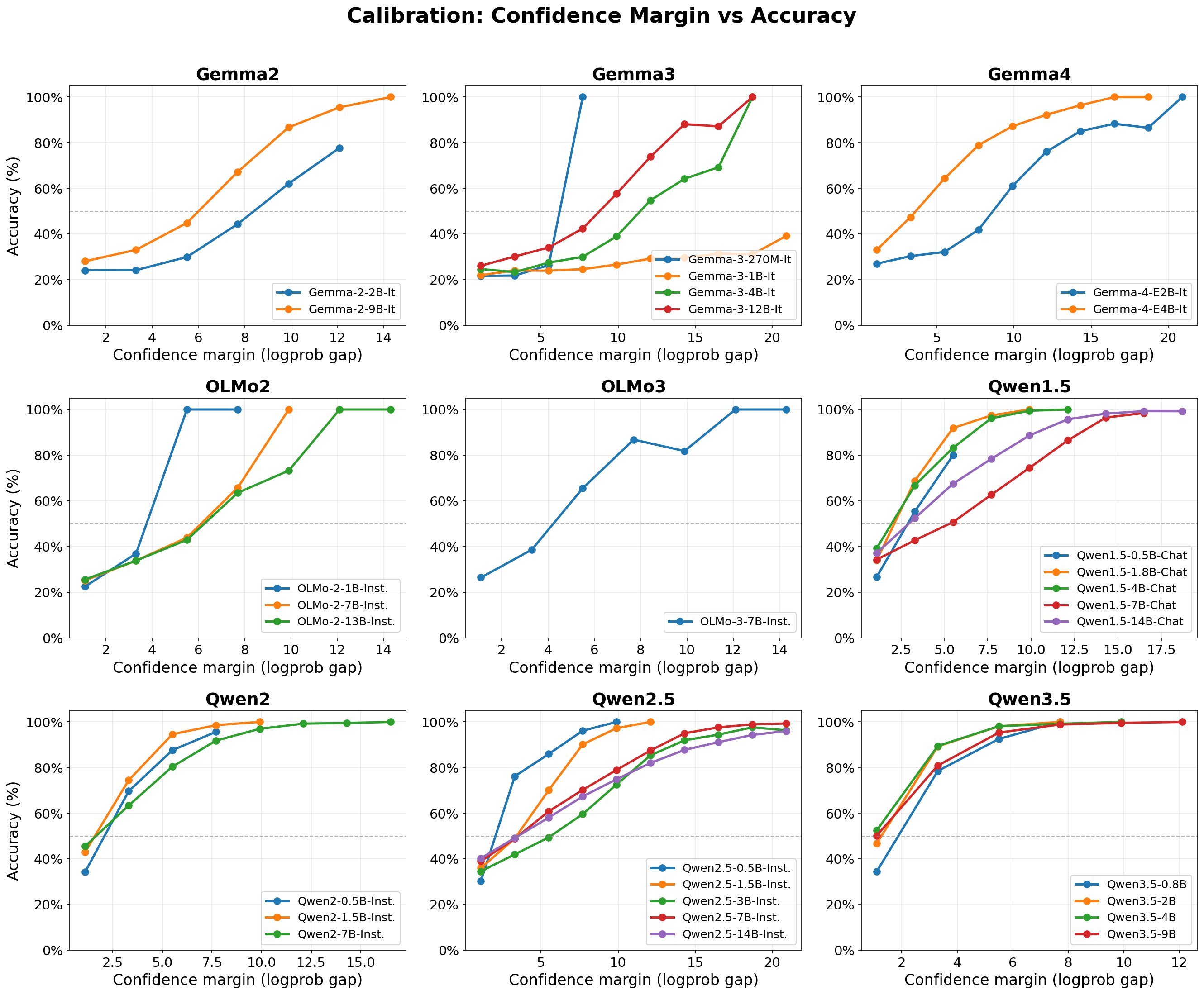}
\caption{\textbf{Accuracy by confidence margin.} Accuracy grouped by log-probability margin, one panel per family. More confidence really does mean more accuracy in every family, but the useful range varies by close to a factor of ten, so margins from different models are not comparable. The panels have independent $x$-axes, so compare the ranges rather than the slopes. Qwen3.5 is the narrowest, fitting its whole curve into a range where the other families are still climbing.}
\label{fig:calibration}
\end{figure}

\begin{figure}[htbp]
\centering
\includegraphics[width=\linewidth]{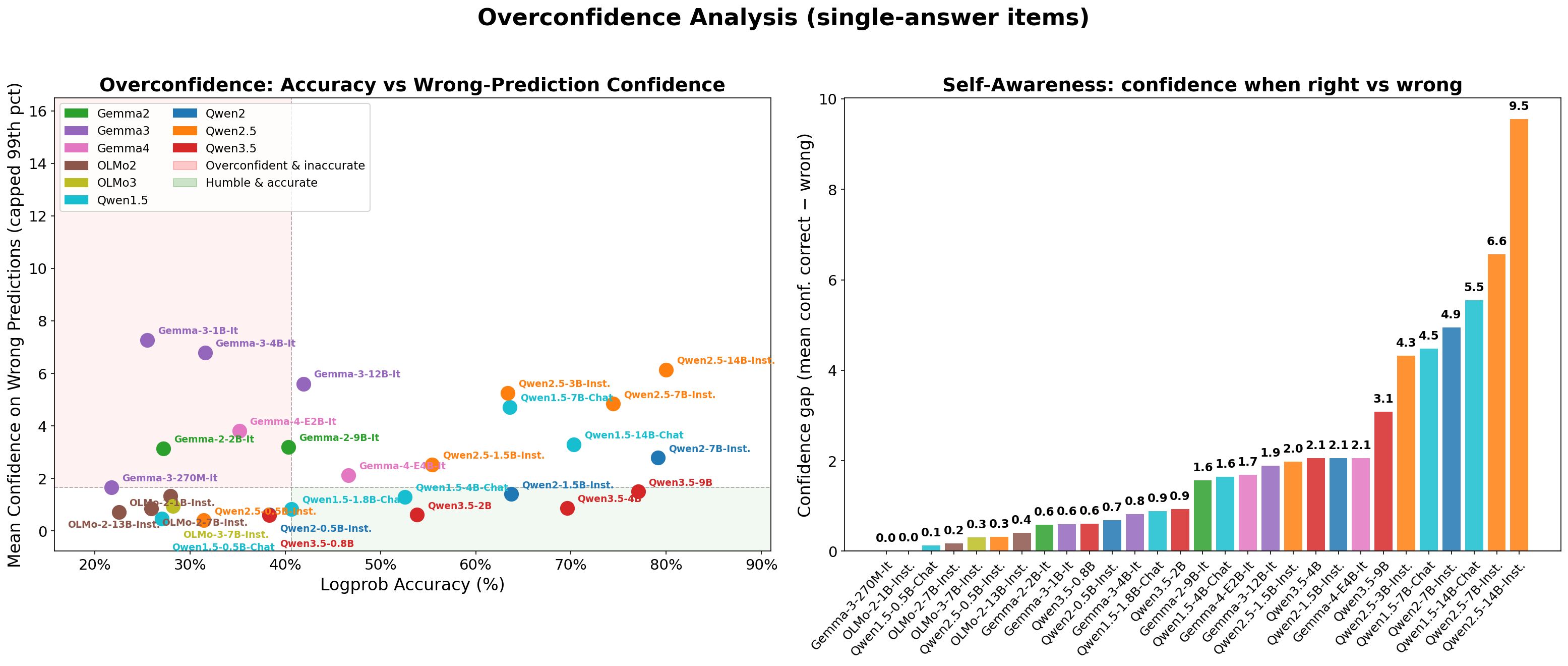}
\caption{\textbf{Overconfidence.} Left: accuracy against average confidence on \emph{wrong} answers, split at the medians. Right: the confidence gap between right and wrong answers, with confidence capped at the 99th percentile. The top-left corner is where models are inaccurate and sure of themselves, and two Gemma3 models sit there. The right panel shows this does not follow from accuracy: Qwen2.5-14B-Inst.\ and Qwen3.5-9B score similarly but separate right from wrong answers by $9.5$ and $3.1$.}
\label{fig:overconfidence}
\end{figure}

\paragraph{Setup.}
We take the margin $m$ of Equation~\eqref{eq:lp} as the confidence signal.
For each model we group its items by margin and measure accuracy within each group, which gives one calibration curve per model (Figure~\ref{fig:calibration}).
We then reduce each model to a single number, the gap between its mean confidence on correct answers and on incorrect ones, with confidence capped at the 99th percentile to keep a few extreme margins from dominating (Figure~\ref{fig:overconfidence}).
A large gap means the model's confidence is worth listening to; a gap near zero means it is not.

\paragraph{Results.}
Two findings stand out.
First, confidence points the right way in every family: accuracy rises with the margin every time.
What changes is the scale.
Qwen2.5 and Gemma4 spread out over margins as large as 20 nats and get nearly everything right at the top end, while Qwen3.5 squeezes its whole curve into a much narrower range, so a margin of $3$ means different things for different models.
Second, calibration does not follow from accuracy.
Qwen2.5-14B-Inst.\ reaches a confidence gap of $9.5$, while Qwen3.5-9B manages only $3.1$ at similar accuracy, and Gemma-3-1B-It and OLMo-2-13B-Inst.\ fall below $1.0$.
Gemma-3-1B-It and Gemma-3-4B-It are the worst case: inaccurate and confident all the same.
A model with a small gap should not be trusted to decide when to answer, however good its overall score is.

\end{document}